\documentclass{article}
\usepackage{graphicx}

    \PassOptionsToPackage{numbers, compress}{natbib}

    \usepackage[final]{neurips_2024}

\usepackage[utf8]{inputenc} 
\usepackage[T1]{fontenc}    
\usepackage{hyperref}       
\usepackage{url}            
\usepackage{booktabs}       
\usepackage{amsfonts}       
\usepackage{nicefrac}       
\usepackage{microtype}      
\usepackage{xcolor}         

\usepackage{tabularx}
\usepackage{booktabs, multirow} 
\usepackage{soul}

\usepackage{xcolor}
\usepackage{color, colortbl}
\definecolor{verylightgreen}{rgb}{0.9, 1, 0.9}
\definecolor{verylightred}{rgb}{1, 0.9, 0.9}
\definecolor{verylightblue}{rgb}{0.8, 0.9, 1.0}
\definecolor{verylightorange}{rgb}{1.0, 0.8, 0.6}
\definecolor{verylightgray}{rgb}{0.95, 0.95, 0.95}

\usepackage{wrapfig,lipsum,booktabs}

\usepackage{subcaption}

\title{\textsc{GSR-Bench}: A Benchmark for Grounded Spatial Reasoning Evaluation via Multimodal LLMs}

\author{%
  Navid Rajabi\\
  Department of Computer Science\\
  George Mason University\\
  \texttt{nrajabi@gmu.edu} \\
  \And
  Jana Ko{\v{s}}eck{\'a} \\
  Department of Computer Science \\
  George Mason University \\
  \texttt{kosecka@gmu.edu} \\
}

\begin{document}

\maketitle


\vspace{-6mm}
\begin{figure*}[h!]
    \centering
  \includegraphics[scale=0.32]{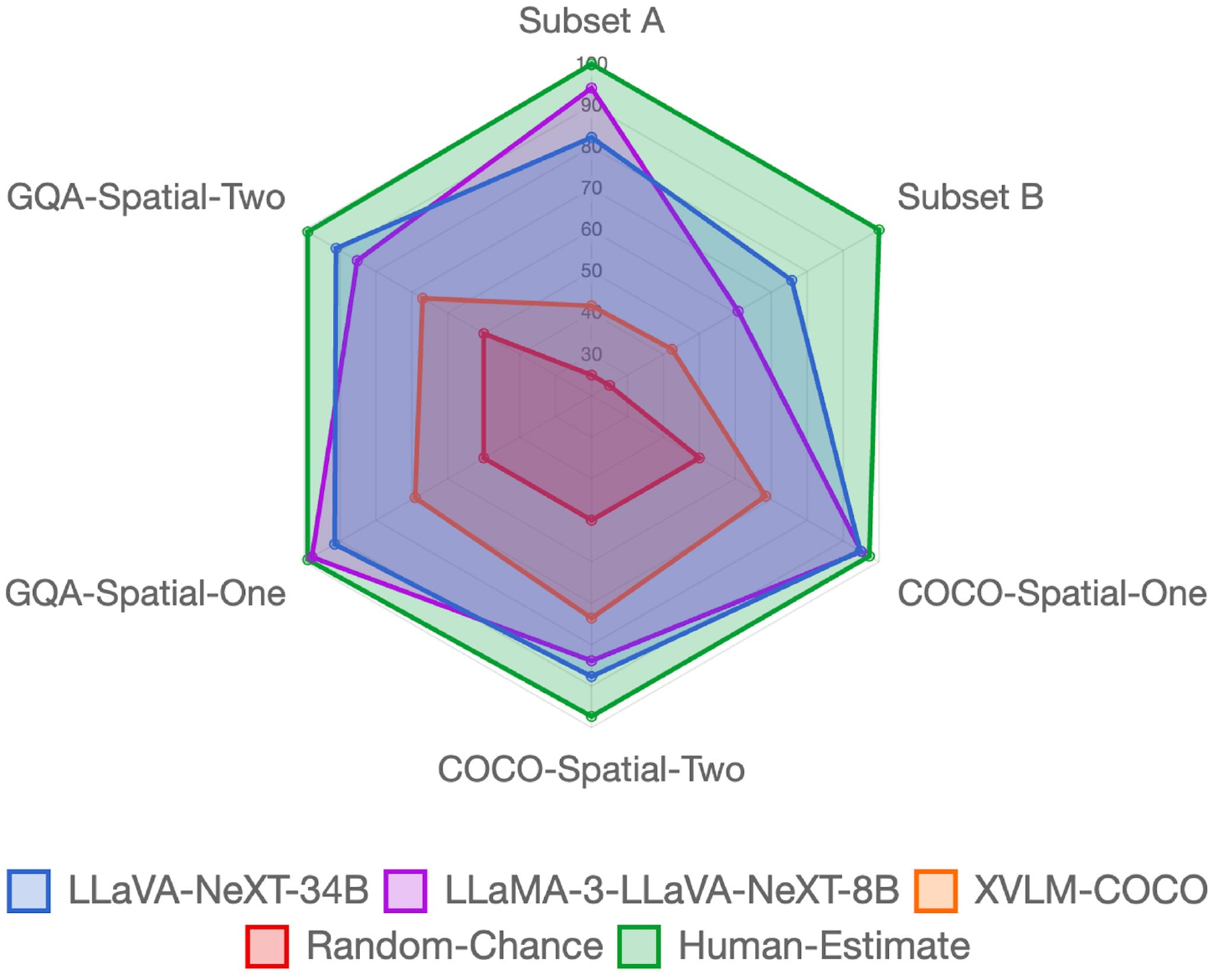}
  \caption {\textsc{LLaMA-3-LLaVA-NeXT-8B} achieves the overall accuracy of 86.1\%, compared to 60.4\% by \textsc{XVLM-COCO}, in What'sUp benchmark, reaching the best trade-off between accuracy and parameters size, since it performs only 1.1\% lower than \textsc{LLaVA-NeXT-34B}, which has $\times$4.25 number of parameters.}
\end{figure*}
\vspace{-4mm}
\begin{abstract}
The ability to understand and reason about spatial relationships between objects in images is an important component of visual reasoning. This skill rests on recognizing and localizing objects of interest and determining their spatial relation. Early vision and language models (VLMs) have been shown to struggle to recognize spatial relations. We extend the previously released What'sUp \cite{kamath-etal-2023-whats} dataset and propose a novel comprehensive evaluation for spatial relationship understanding that highlights the strengths and weaknesses of 9 Multimodal LLMs (MLLMs), in comparison with the 18 VLMs tested in What'sUp dataset. Our experiments encompass three classes of MLLMs that vary in their parameter sizes (ranging from 7B to 110B), training/instruction-tuning methods, and visual resolution to benchmark their performances and scrutinize the scaling laws in this task. 
\end{abstract}
\vspace{-6mm}
\section{Introduction}
\vspace{-2mm}
\begin{figure*}[!h]
\centering
  \includegraphics[width=1\linewidth]{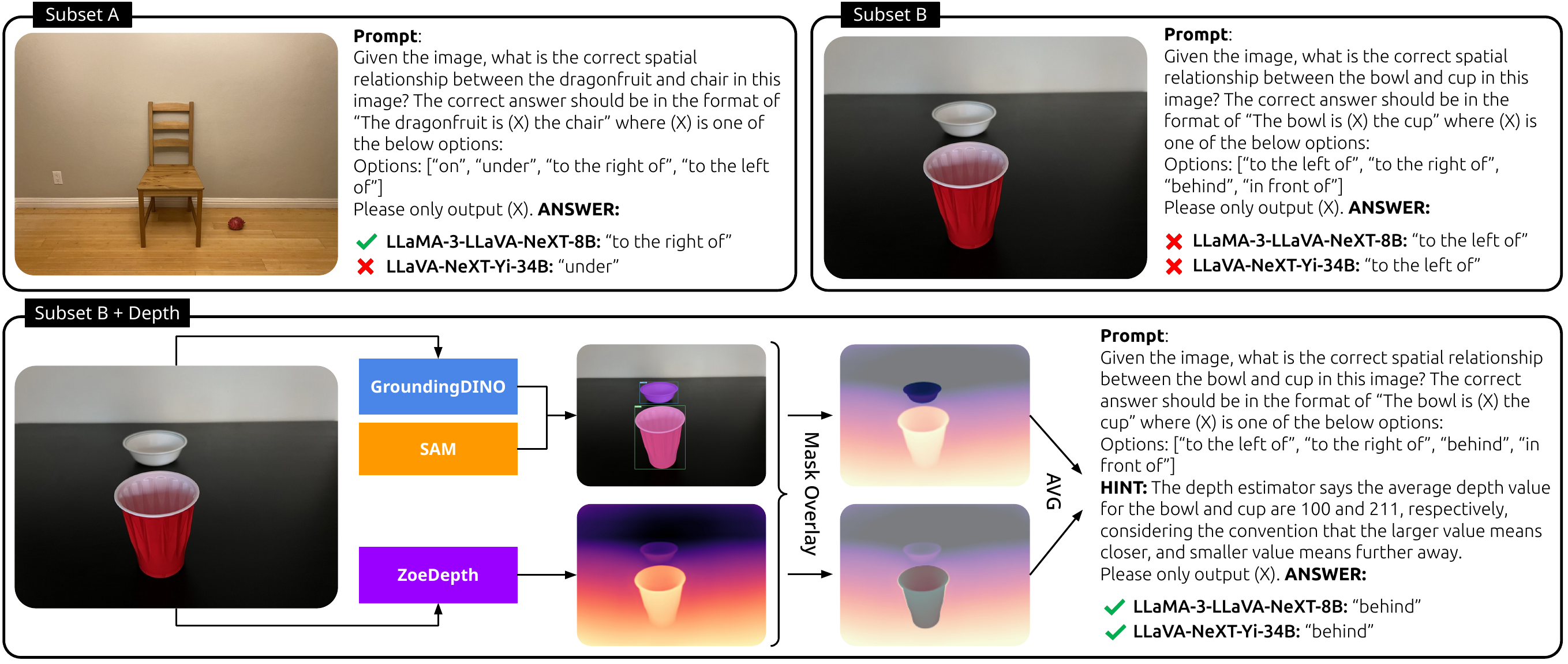}
  \caption {Our pipeline overview for spatial relationship understanding prompting, shown in the top two figures, and our depth-augmented prompting, shown in the bottom figure.}
  \label{fig:main_pipeline}
\end{figure*}
Earlier efforts for benchmarking vision and language models (VLMs) were developed for cross-modal and/or dual-encoder, end-to-end models, like LXMERT \cite{tan2019lxmert}, CLIP \cite{clip}, BLIP \cite{blip}, with the focus on downstream tasks performances such as VQA \cite{vqa}, GQA \cite{hudson2019gqa}, referring expressions \cite{referitgame}, image-text matching or image/text retrieval.
While spatial relations are often part of VQA datasets, the evaluation of spatial reasoning is often conflated with grounding referring expressions or objects and their attributes\footnote{VQA example question may be: “Is there a woman to the left of the person that is wearing a wetsuit?”}. To isolate these issues, authors in~\cite{kamath-etal-2023-whats} introduced a new benchmark that focuses on spatial relationship understanding only. 
Using image-text matching evaluation methodology, they showed that contrastive models
such as CLIP, BLIP, and their follow-up variants struggle to understand spatial relations with the
best accuracy around $61$\%. Recent advances in generative large multi-modal models have shown remarkable visual knowledge and reasoning 
capabilities. We revisit the spatial relationship understanding in the context of MLLMs and extend the existing What'sUp benchmark~\cite{kamath-etal-2023-whats} to include bounding box annotations and depth information. Compositional spatial relationship understanding requires successful recognition of objects and determining their locations. Furthermore, the knowledge of scene depth helps to disambiguate certain relationships (e.g., "\textit{in front of}" or "\textit{behind}").  
The availability of this information can support a grounded understanding of spatial relations and 
will contribute to the fine-grained evaluation of large generative MLLMs, which lag behind their earlier counterparts. A few exceptions are multi-task multi-modal benchmarks like MMBench~\cite{liu2023mmbench} and 
its related benchmarks
that focus on evaluating several MLLMs for both visual recognition tasks and description generation.  
Given the simple structure of spatial clauses, we can study separately the ability of the model to ground the subject and object in the clause, and the effect and means of incorporating the depth information. 
The contributions of this work can be summarized as follows: (1) Extended What'sUp spatial relationship dataset with depth, masks, and bounding box annotations, (2) Design of different prompting strategies through structured prompting for the evaluation of grounding and spatial reasoning, and (3) Comprehensive evaluation and comparison of 18 VLMs and 9 MLLMs, with various sizes, resolutions, pre-training/instruction-tuning, and prompting strategies.

\begin{table*}[t]
\centering
\scriptsize
\resizebox{\textwidth}{!}{  
\begin{tabular}{l|c|c|c|cc|cc|c}
\toprule                        \textbf{\textsc{Model}}
                                & \multicolumn{1}{c}{\textbf{\textsc{Num}}}
                                & \multicolumn{1}{c}{\textbf{\textsc{Subset A}}} & \multicolumn{1}{c}{\textbf{\textsc{Subset B}}} & \multicolumn{2}{c}{\textbf{\textsc{COCO-spatial}}} & \multicolumn{2}{c}{\textbf{\textsc{GQA-spatial}}} & \multirow{2}{*}{\begin{tabular}[c]{@{}c@{}}\textbf{\textsc{Total}}\\ \textsc{Average}\end{tabular}} \\
                                & \textsc{Params} & \textsc{Sub-Obj} & \textsc{Sub-Obj} & \textsc{One-Obj} & \textsc{Two-Obj}  & \textsc{One-Obj}  & \textsc{Two-Obj}  &                               \\
                                \midrule

CLIP ViT-B/32 \scriptsize{\cite{clip}} & 151M & 30.3 & 31.6 & 43.7 & 51.1 & 46.5 & 47.4 & 41.8 \\

CLIP ViT-L/14 & 428M & 26.5 & 25.7 & 49.2 & 49.8 & 46.1 & 48.5 & 41.0 \\

NegCLIP \scriptsize{\cite{bag-of-words}} & -- & 32.5 & 36.3 & 47.4 & 46.4 & 45.3 & 46.7 & 42.4 \\

RoBERTaCLIP \scriptsize{\cite{kamath-etal-2023-whats}} & -- & 25.2 & 25.0 & 46.3 & 53.6 & 50.8 & 48.8 & 41.6 \\

CoCa \scriptsize{\cite{coca}} & 2.1B & 29.4 & 29.4 & 48.1 & 45.2 & 45.0 & 49.1 & 41.0 \\

XVLM 4M \scriptsize{\cite{xvlm}} & 216M & 40.0 & 23.0 & 58.4 & 65.0 & 62.8 & 54.6 & 50.6 \\

XVLM 16M & 216M & 50.7 & 33.1 & 65.4 & 64.5 & 63.2 & 53.3 & 55.0 \\

BLIP 14M \scriptsize{\cite{blip}} & 583M & 38.8 & 38.2 & 54.2 & 53.9 & 49.1 & 50.5 & 47.5 \\

BLIP 129M & 583M & 30.3 & 30.4 & 44.8 & 53.9 & 50.5 & 47.4 & 42.9 \\

BLIP2-ITM \scriptsize{\cite{blip2}} & 188M & 44.9 & 30.4 & 48.3 & 57.7 & 46.0 & 53.6 & 46.8 \\

BLIP2-ITC & 188M & 35.9 & 22.1 & 55.6 & 51.8 & 52.6 & 49.5 & 44.6 \\

FLAVA \scriptsize{\cite{flava}} & -- & 33.7 & 27.2 & 50.3 & 55.0 & 52.2 & 51.2 & 44.9 \\

\midrule

CoCa-Caption & 2.1B & 25.5 & 22.8 & 45.9 & 51.4 & 48.5 & 50.5 & 40.8 \\

XVLM-Flickr30K & 216M & 45.1 & \underline{\textit{43.4}} & 63.1 & 67.3 & 64.7 & 58.1 & 56.9 \\

XVLM-COCO & 216M & 41.7 & 42.4 & \underline{\textit{68.4}} & \underline{\textit{73.6}} & \underline{\textit{69.1}} & \underline{\textit{67.0}} & \underline{\textit{60.4}} \\

BLIP-Flickr30K & 583M & 29.6 & 38.0 & 50.0 & 58.4 & 50.3 & 47.4 & 45.6 \\

BLIP-COCO & 583M & 35.7 & 29.9 & 46.4 & 56.4 & 50.3 & 52.6 & 45.2 \\

BLIP-VQA & 583M & \underline{\textit{57.8}} & 37.7 & 63.6 & 60.5 & 63.8 & 52.9 & 56.0 \\

\midrule
\rowcolor{verylightred}
\textsc{LLaVA-1.5-Vicuna} & 7B & 25.0 & 31.9 & 90.4 & 66.6 & 91.2 & 62.9 & 61.3 \\
\rowcolor{verylightred}
\textsc{LLaVA-1.5-Vicuna}  & 13B & 58.5 & 28.2 & 92.5 & 78.9 & 93.1 & 82.8 & 72.3 \\
\midrule
\rowcolor{verylightgreen}
\textsc{LLaVA-NeXT-Mistral} & 7B & 37.4 & 22.0 & 81.1 & 60.4 & 89.4 & 57.0 & 57.9 \\

\rowcolor{verylightgreen}
\textsc{LLaVA-NeXT-Vicuna}  & 7B & 38.6 & 26.2 & \underline{95.5} & 71.8 & \underline{97.6} & 79.0 & 68.1 \\

\rowcolor{verylightgreen}
\textsc{LLaVA-NeXT-Vicuna}  & 13B & 75.0 & 20.1 & \textbf{95.6} & 78.6 & \underline{97.6} & 84.9 & 75.3 \\

\rowcolor{verylightgreen}
\textsc{LLaMA-3-LLaVA-NeXT}  & 8B & \textbf{94.2} & 60.8 & 95.1 & 83.9 & \textbf{97.8} & 85.2 & \underline{86.1} \\

\rowcolor{verylightgreen}
\textsc{LLaVA-NeXT-Yi} & 34B & 82.3 & \textbf{75.7} & 94.8 & \textbf{87.7} & 91.5 & \underline{91.1} & \textbf{87.2} \\

\rowcolor{verylightgreen}
\textsc{LLaVA-NeXT-Qwen1.5} & 110B & \underline{93.9} & 54.2 & 90.6 & \underline{84.1} & 96.2 & \textbf{94.2} & 85.4 \\
\midrule
\rowcolor{verylightblue}
\textsc{Intern-VL-Chat-1.5} & 26B & 92.2 & \underline{61.8} & 95.1 & 82.3 & \textbf{97.8} & 82.8 & 85.3 \\
\midrule
Random Chance  & -- & 25.0 & 25.0 & 50.0 & 50.0 & 50.0 & 50.0 & 41.7 \\

\bottomrule
\end{tabular}
}
\caption{Template-based generation (TG) results using CircularEval. The first two sections come from What'sUp \cite{kamath-etal-2023-whats} results. The rest shows our LLaVA \colorbox{verylightred}{1.5}, \colorbox{verylightgreen}{1.6}, and \colorbox{verylightblue}{Intern-VL-1.5} prompting results. Our best-performing is \textbf{bold}, 2nd-best \underline{underlined}, and What'sUp best-performing \underline{\textit{italic underlined}}.
}
\label{tab:tg_results_full}
\end{table*}
\vspace{-4mm}
\section{GSR Benchmark}
\vspace{-3mm}
We extend carefully curated What'sUp dataset~\cite{kamath-etal-2023-whats} that is comprised of Subset A containing pairs of objects in unambiguous spatial relations, being "\textit{on}", "\textit{under}", "\textit{left of}" or "\textit{right of}" a table, chair, or armchair, and Subset B containing an object "\textit{in front of}", "\textit{behind}", "\textit{left}" or "\textit{right}" of another object on a tabletop, and subsets of COCO-Spatial and GQA-Spatial with either one or two objects occurring, accompanied by spatial clauses like "\textit{on top of}", "\textit{on the bottom of}", "\textit{right of}", or "\textit{left of}".
To study the grounding in this context, we annotate the dataset with bounding box coordinates and segmentation masks for all the objects mentioned in the captions and the depth maps for the images.  We leverage GroundingDINO \cite{liu2023groundingdino} as an open-vocabulary object detector, Segment Anything (SAM) \cite{kirillov2023sam} for the object mask segmentation, and ZoeDepth \cite{bhat2023zoedepth} for monocular depth estimation. In the next section, we explain in detail how these additional annotations enable a more rigorous and grounded evaluation of spatial reasoning and its components\footnote{All the code and data will be publicly available.}.

\vspace{-3mm}
\section{\textsc{GSR-Bench} Experiments}
\vspace{-3mm}
Grounded spatial reasoning evaluation is typically done using 
image-text matching, binary VQA, or multiple-choice VQA. Further evaluations 
include subject and/or object grounding and localization; and exploring the effect of using depth information. In addition to 18 VLMs that have been evaluated in~\cite{kamath-etal-2023-whats}, we focus on the probing of 
open-source generative MLLMs like LLaVA and Intern-VL\footnote{Intern-VL is the leading model in MMBench.} using structured generation methodologies of Multiple choice (MC) and Template-based generation (TG).
In MC prompting, captions for each image are represented as A, B, C, and D options for Subset A and Subset B, while A and B options for COCO-Spatial and GQA-Spatial Subsets. Then, the model is prompted to choose the correct letter as the final answer.
\begin{table*}[!h]
\centering
\scriptsize
\resizebox{\textwidth}{!}{  
\begin{tabular}{l|cc|cc|c|cc|c|cc|c}
\toprule
\textbf{\textsc{Model}}
& \multicolumn{2}{c}{\textbf{\textsc{Subset A}}} & \multicolumn{2}{c}{\textbf{\textsc{Subset B}}} & \multicolumn{3}{c}{\textbf{\textsc{COCO-spatial}}} & \multicolumn{3}{c}{\textbf{\textsc{GQA-spatial}}} & \multirow{2}{*}{\begin{tabular}[c]{@{}c@{}}\textbf{\textsc{Avg}}\\ \textsc{G-Score}\end{tabular}} \\
& \textsc{Sub} & \textsc{Obj} & \textsc{Sub} & \textsc{Obj} & \textsc{One-Obj} & \textsc{Sub} & \textsc{Obj} & \textsc{One-Obj} & \textsc{Sub} & \textsc{Obj} & \\

\midrule

\textsc{LLaVA-1.5-Vicuna-7B} & 9.7 & 79.4 & 51.5 & 25.7 & 47.4 & 49.8 & 48.0 & 31.9 & 55.0 & 47.8 & 44.62 \\

\textsc{LLaVA-1.5-Vicuna-13B} & 13.8 & 86.1 & 77.4 & 32.3 & 60.9 & 61.8 & 61.0 & 42.1 & 72.2 & 59.8 & 56.74 \\

\midrule

\textsc{LLaVA-NeXT-Vicuna-7B} & 14.1 & 99.0 & 95.8 & 66.7 & 81.9 & 84.5 & 77.7 & 45.5 & 60.1 & 56.0 & 68.13 \\

\textsc{LLaVA-NeXT-Mistral-7B} & 13.1 & 82.3 & 93.9 & 60.0 & \underline{87.1} & 86.8 & \underline{85.7} & 69.2 & 85.9 & 81.8 & 74.58 \\
\textsc{LLaVA-NeXT-Vicuna-13B} & 15.3 & 84.0 & 95.3 & 67.6 & \underline{87.1} & \textbf{90.2} & 83.9 & 69.6 & 85.9 & 80.4 & 75.93 \\

\textsc{LLaMA-3-LLaVA-NeXT-8B}   & 19.2 & \underline{99.3}  & 96.6 & 73.8 & 85.7 & 87.5 & 83.2 & 69.0 & 84.5 & 80.4 & 77.92 \\

\textsc{LLaVA-NeXT-Yi-34B} & \underline{21.1} & \textbf{100.0} & \underline{97.8} & \underline{78.9} & 83.7 & 85.7 & 81.4 & \underline{70.0} & \textbf{88.0} & \underline{83.5} & \underline{79.01} \\

\textsc{LLaVA-NeXT-Qwen1.5-110B} & \textbf{29.4} & 98.5 & \textbf{98.8} & \textbf{80.1} & \textbf{88.7} & \underline{88.2} & \textbf{86.4} & \textbf{74.9} & \underline{86.9} & \textbf{84.2} & \textbf{81.61} \\

\midrule
\rowcolor{verylightgray}
GroundingDINO 
[avg($\rho$)] & 58.8 & 92.0 & 78.1 & 70.1 & 62.3 & 62.8 & 59.3 & 59.4 & 70.4 & 65.2 & 67.84 \\

\rowcolor{verylightgray}
GroundingDINO 
[$\Sigma$($\rho$ $\geq$ 0.5)/$t$] & 68.9 & 100.0 & 90.0 & 88.7 & 71.0 & 73.6 & 66.4 & 59.1 & 76.3 & 71.1 & 76.51 \\

\bottomrule
\end{tabular}
}
\caption{Grounding/Localization results. \textsc{AVG G-Score} refers to the mean accuracy of IoU $\geq$ 0.5. The bottom two rows refer to the GroundingDINO mean confidence scores ($\rho$), and mean accuracy of $\rho$ $\geq$ 0.5, respectively. 
}
\label{tab:auto_annot_grounding_pipeline}
\end{table*}
We ran each prompt with 4 different permutations so as to vary the position of the answer among the choices in MC and the list of options in TG prompting. An instance is considered correct if all four options are predicted correctly, known as \textit{CircularEval}, introduced in MMBench \cite{liu2023mmbench}. As opposed to the CircularEval, there exists VanillaEval, which only asks the model to choose the correct answer from a list of options once and has been shown to be prone to bias in recent studies. We first ran our experiments using MC prompting and observed a significant degree of bias among the models when the position of the answer varied among the choices of A, B, C, or D. This bias and sensitivity turned out to be even more detrimental in smaller models, while larger models like \textsc{LLaVA-NeXT-Yi-34B} and \textsc{LLaMA-3-LLaVA-NeXT-8B} showed significantly higher robustness (See Figure \ref{fig:mc_barcharts} in the Appendix for details). 
This phenomenon also corroborates the findings of multiple recent studies in LLMs \cite{zheng2023llmsnotrobustmc, pezeshkpour2023largemodelssensitivity, wang2023llmsnotfairevals, Xue2024StrengthenedSBMC, wang2024lookatthetext}.
According to this observation, we opted for TG prompting, accompanied by the CircularEval methodology, inspired by Gemini 1.5 Pro \cite{reid2024gemini}. 
See Table \ref{tab:tg_results_full} for the TG prompting results, where rows in section 1 and 2 come from the What'sUp benchmark~\cite{kamath-etal-2023-whats}, section 3 refers to LLaVA-1.5 models~\cite{liu2023llava}, section 4 to the LLaVA-NeXT models~\cite{li2024llavanext-strong, liu2024llavanext}, and section 5 to the InternVL-1.5 results~\cite{chen2024finternVL}.
\vspace{-3mm}
\paragraph{Grounding/Localization Evaluation.}
This experiment aims to measure the MLLMs grounding ability of the objects mentioned in the captions. 
 Recent studies like \cite{rajabi2023towards} on Visual Spatial Reasoning (VSR) benchmark \cite{liu2023vsrTACL} has demonstrated that there exist multiple cases where the VLM correctly predicts the binary ITM label of 1 using the holistic representations of the image and caption, while the model fails to localize the subject and object correctly.
Our experiments aim to quantify these type of behaviors in MLLMs. 
We prompt MLLMs to extract the normalized bounding box coordinates for the caption's objects as "\texttt{Give me the bounding box coordinates for the \{object\}}"
and compute the IoU between the model's output and the GroundingDINO output for each object, assigning the binary accuracy of 1 if IoU $\geq$ 0.5, otherwise 0. See Table \ref{tab:auto_annot_grounding_pipeline} for the results. 



\vspace{-2mm}
\begin{wraptable}{r}{6cm}
\scriptsize
\begin{tabular}{lcc}
\toprule
\multicolumn{1}{c}{\textsc{Model}} & \multicolumn{1}{c}{\begin{tabular}[c]{@{}c@{}}\textbf{\textsc{w/o}}\\ \textsc{depth} 
\end{tabular}} & \multicolumn{1}{c}{\begin{tabular}[c]{@{}c@{}} \textbf{\textsc{with}} \\ \textsc{depth} \end{tabular}} \\
\midrule
\textsc{InternVL-Chat-1.5-26B}  & 26.5  & 40.7 \\
\textsc{LLaMA-3-LLaVA-NeXT-8B}  & 53.4  & 60.3 \\
\textsc{LLaVA-NeXT-Yi-34B} & 64.7  & 81.9 \\
\bottomrule
\end{tabular}
\caption{DAP results for \textit{behind} \& \textit{in front of} cases.}
\label{tab:dap_experiment_results}
\end{wraptable} 
\vspace{-1mm}
\paragraph{Depth-Augmented Prompting (DAP).} 
The experiments in Table~\ref{tab:tg_results_full} revealed that Subset B is the lowest-performing, 
with many instances requiring reasoning about "\textit{behind}" and "\textit{in front of}" spatial clauses. 
We propose to incorporate the depth values of \texttt{subject} and \texttt{object} into the prompt, as a hint to the model, utilizing our augmented benchmark annotations, depicted in Figure \ref{fig:main_pipeline}. We show that this minimal change improves the accuracy of top-3 performing models in these instances of Subset B, reported by CircularEval in Table \ref{tab:dap_experiment_results}. 
\vspace{-3mm}
\paragraph{Discussion.}
According to Table \ref{tab:tg_results_full} and \ref{tab:auto_annot_grounding_pipeline}, there is a positive correlation, even stronger in grounding, between \textbf{scaling the LLM size} \& \textbf{visual resolution}, and the \textbf{overall accuracy} in both tasks. Conversely, there exist multiple exceptions, which are inevitable to concretely justify due to various intervening factors, such as (1) differences in training/fine-tuning \& architectures and (2)
release date and further instruction-tuning of the LLMs, like \textsc{LLaMA-3-8B}, which has the most-recent knowledge cut-off. Grounding small objects, which refers to the \textsc{Sub} column in Subset A, seems challenging for all, and worst in smaller models, according to Table \ref{tab:auto_annot_grounding_pipeline}. 
We also observed a plateau in 
Table \ref{tab:tg_results_full}, especially in \textsc{Qwen-1.5-110B}, which is the largest ever released open-source MLLM at the moment.
This could be a sign of saturation in reasoning while grounding is still going up by scaling.
Since the lowest grounding results have been achieved in localizing the subjects in Subset A, we also ran an additional Causal Effect analysis using linear regression to measure the causality between the grounding accuracy, as the independent variable, and the reasoning accuracy, as the dependent variable. As a result, we could reject the null hypothesis that "There is no causal effect between grounding and reasoning" due to the regression coefficient of $\beta$ = 6.025 and the $\rho$-value of 0.007, which is significantly lower than 0.05, shown in Figure \ref{fig:causaleffect}.
\vspace{-4mm}
\section{Conclusions}
\vspace{-3mm}
In this work, we introduce a new benchmark for grounded spatial reasoning by enriching the What'sUp dataset with additional supervision for a more fine-grained assessment of MLLM's spatial understanding. We also propose a new compositional evaluation methodology for (1) a stricter assessment 
of spatial relationship understanding through CircularEval, 
and (2) measuring the model's grounding capability using the labels we generate through our cost-effective auto-annotation pipeline. Our evaluations reveal the superiority of LLaVA MLLMs over the best-performing VLMs evaluated in What'sUp, like XVLM, by a significant margin of $\sim +$26.8\%.
Future works may investigate the remaining gap between the top open-source MLLMs and human-level accuracy.

{\small
\bibliographystyle{natbib}
\bibliography{neurips_2024}
}





\newpage
\appendix

\section{Appendix / supplemental material}

\subsection{Limitations}
\paragraph{Small-scale Dataset.} Our split sizes remain the same as the What'sUp dataset in which Subset A has 412, Subset B has 408, COCO-Spatial-One has 2247, COCO-Spatial-Two has 440, GQA-Spatial-One has 1160, and GQA-Spatial-Two has 291 instances. Although this benchmark includes 4,958 image instances in total, each instance covering one or two objects, with various domain shifts in each 6 split, it is smaller than already existing benchmarks related to spatial reasoning, like Visual Genome \cite{krishna2017visualgenome}, GQA \cite{hudson2019gqa}, VSR \cite{vsr}, SpatialSense \cite{yang2019spatialsense}, MMBench \cite{liu2023mmbench}, etc. The reason is that this work aims to provide a carefully curated benchmark for spatial relationship understanding evaluation in a controlled setting to abstract away intervening factors that make the evaluations noisy.   

\paragraph{Limited Spatial Prepositions.}
Following the What'sUp dataset, our benchmark is also confined to the primitive spatial clauses of \textit{on}, \textit{under}, \textit{behind}, \textit{in front of}, \textit{to the left of}, \textit{to the right of}, \textit{below} and \textit{above}, when having two objects involved in the caption, and, \textit{on the top}, \textit{on the bottom}, \textit{on the left} and \textit{on the right} when having only one object in the caption, like in COCO-Spatial-One and GQA-Spatial-One.

\paragraph{Lack of Robustness in MC Prompting.}
In addition to the similar findings of MC noisiness in LLMs that we discussed earlier, we hypothesize that the higher degree of variance in multiple-choice results in the last two subsets (COCO and GQA), which is more significant in the smaller models, could be due to the language domain distribution shift. Most of the LLMs and MLLMs are being trained and evaluated with 4 options in the multiple-choice settings. Conversely, in the last two subsets, we have two captions per image, which means we only provide options A and B to the model in the prompt instead of ABCD without any fine-tuning for this task or this specific type of prompting.

\paragraph{Intern-VL-1.5 Poor Grounding Observation.} An unexpected, significant noisiness in the output of grounding/localization prompting of InternVL-1.5 model prevented us from analyzing and reporting the results for this model, which requires further investigation since a similar behavior has been observed through our interaction with the InternVL-1.5 demo, as well.

\paragraph{Depth Augmentation Nuances.} The issue we noticed in the DAP experiment was the distraction the depth hint can cause in cases where multiple correct relationships hold in the image. For instance, object A can be \textit{to the left of} object B, and also \textit{in front of} object B, at the same time. So, in these ambiguous cases, incorporating depth could make the model's decision biased towards the \textit{in front of} preposition, while the ground-truth might be \textit{to the left of} in this case. Therefore, we believe that trying both prompts, with and w/o depth hint, would be helpful for disambiguation in such cases.

\paragraph{No Human Annotation.} Due to the resource constraints, our extended benchmark relies on the pseudo-labels we generate using state-of-the-art, off-the-shelf models like GroundingDINO, SAM, and ZoeDepth. Future works could incorporate human inspection and labeling for further robustness in annotations. 

\subsection{Computational Resources}
For LLaVA-1.5 (7B and 13B) and LLaVA-1.6/LLaVA-NeXT models up to the 13B versions, we used one 40 GB NVIDIA A100. For LLaVA-Yi-34B, and Intern-VL-Chat-1.5 26B, we used one 80 GB NVIDIA A100. For LLaVA-NeXT-Qwen-1.5 110B, we used four 80 GB NVIDIA A100 GPUs. In most cases, except the Intern-VL model, we used the Huggingface \cite{wolf2019huggingface} library setup and checkpoints.

\begin{figure*}[!h]
  \includegraphics[width=1\linewidth]{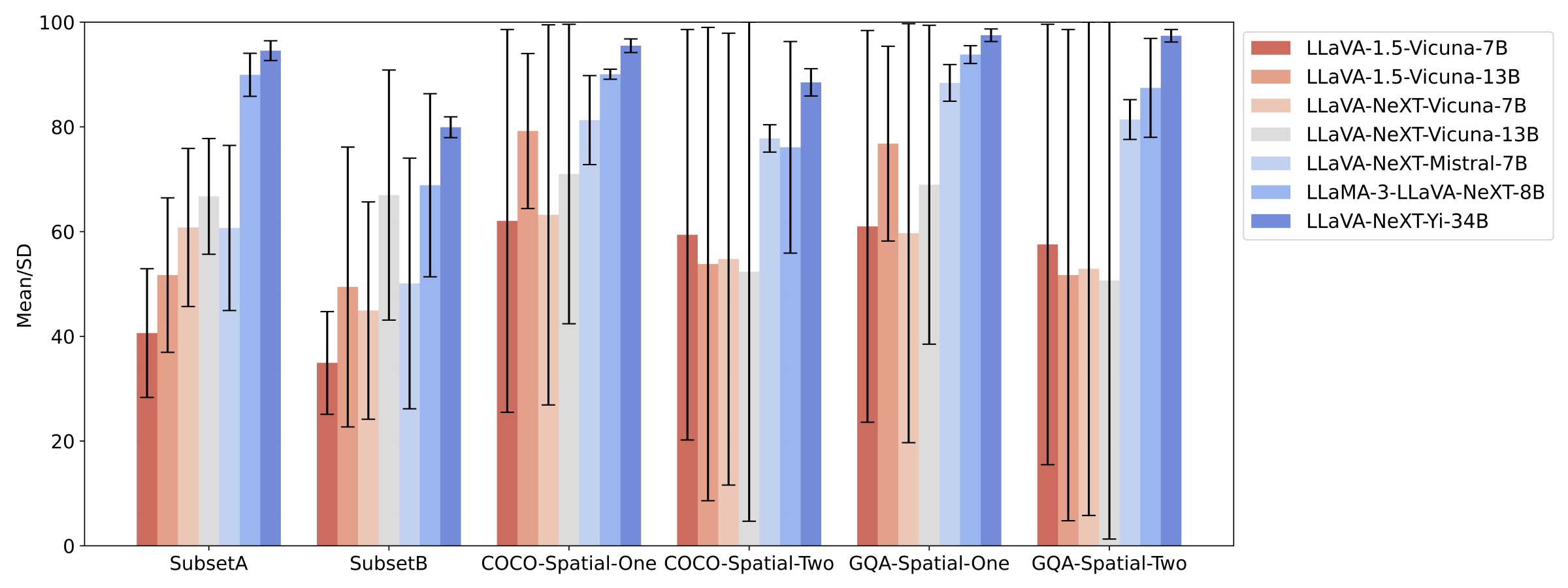}
  \caption {Sensitivity of the models to different permutations of choice order, in the multiple-choice (MC) experiment, which is more significant in the smaller models, and when having two choices of A and B instead of regular 4-choice of A, B, C, and D. \textsc{LLaVA-NeXT-Yi-34B} demonstrates an excellent robustness against this issue.}
  \label{fig:mc_barcharts}
\end{figure*}


\begin{figure*}[!h]
  \includegraphics[width=1\linewidth]{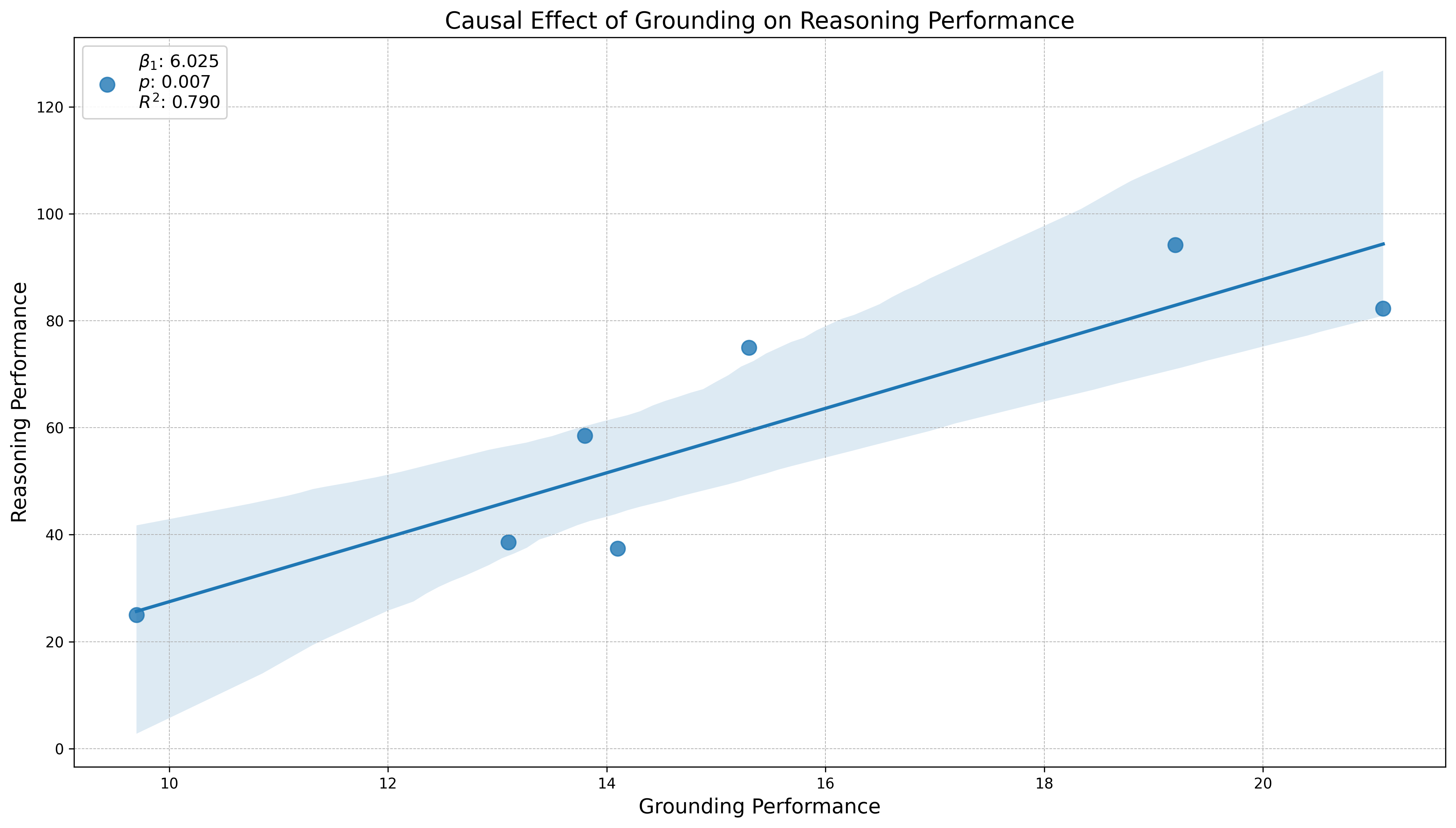}
  \caption {Causal effect analysis between grounding and reasoning accuracy, for the Subset A subjects that was the most difficult setting for the models for localization.}
  \label{fig:causaleffect}
\end{figure*}


\begin{figure*}[!h]
    \centering
    \begin{subfigure}[b]{0.45\textwidth}
        \centering
        \includegraphics[scale=0.25]{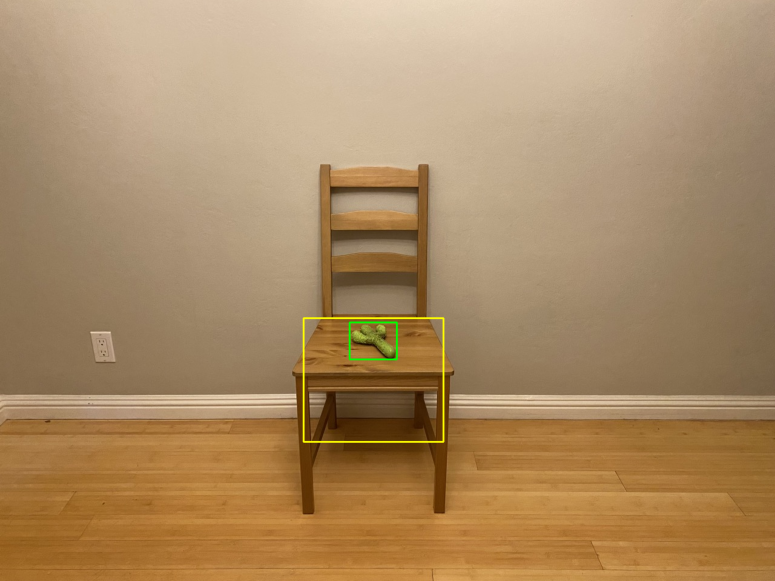}
        \caption{\textbf{toy cactus} on chair}
        \label{fig:sub1}
    \end{subfigure}
    \hfill
    \begin{subfigure}[b]{0.45\textwidth}
        \centering
        \includegraphics[scale=0.25]{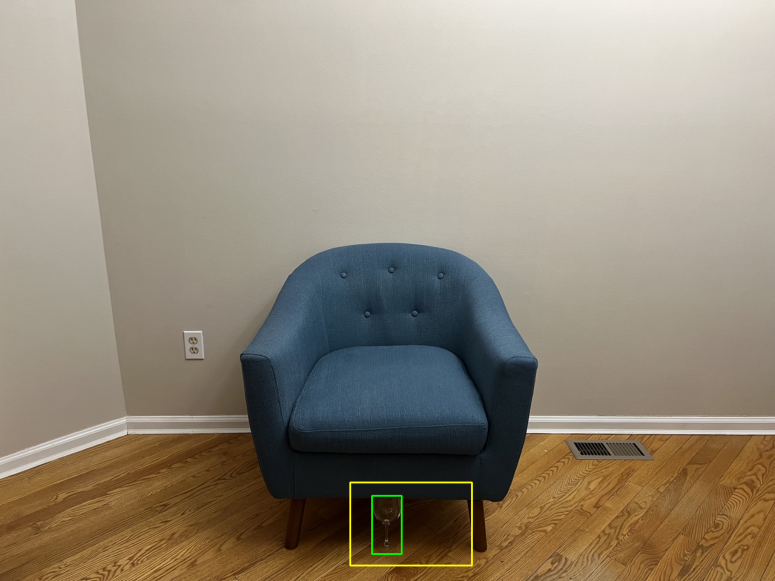}
        \caption{\textbf{wineglass} under armchair}
        \label{fig:sub2}
    \end{subfigure}

    \begin{subfigure}[b]{0.45\textwidth}
        \centering
        \includegraphics[scale=0.25]{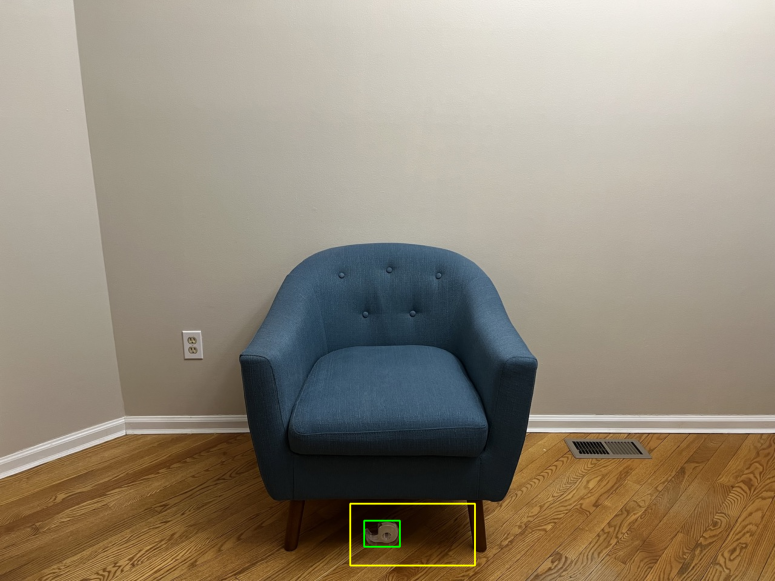}
        \caption{\textbf{tape} under armchair}
        \label{fig:sub3}
    \end{subfigure}
    \hfill
    \begin{subfigure}[b]{0.45\textwidth}
        \centering
        \includegraphics[scale=0.25]{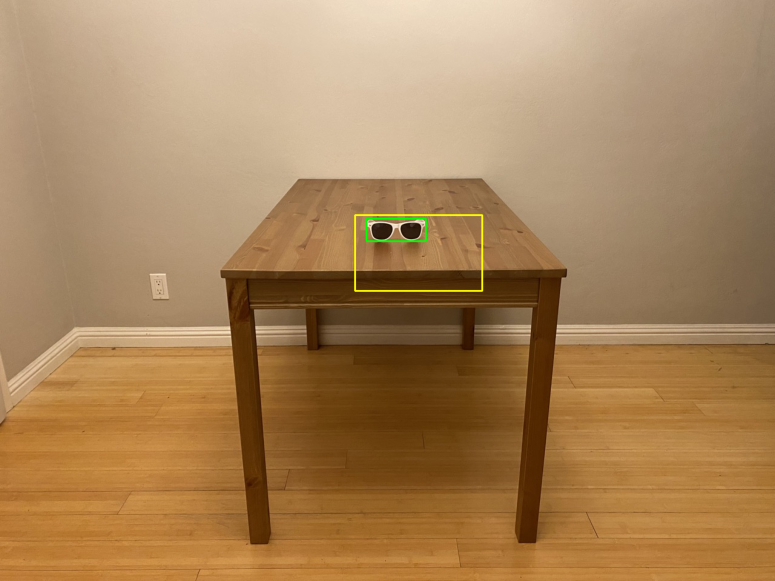}
        \caption{\textbf{sunglasses} on table}
        \label{fig:sub4}
    \end{subfigure}

    \begin{subfigure}[b]{0.45\textwidth}
        \centering
        \includegraphics[scale=0.25]{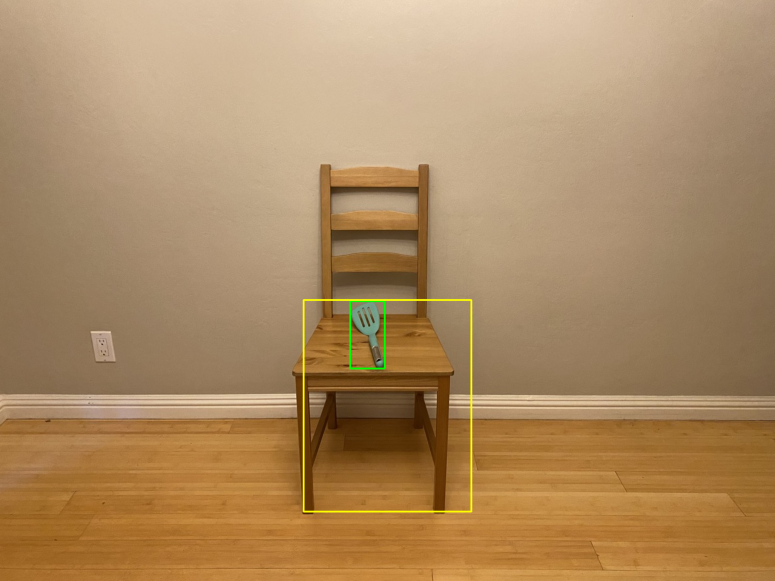}
        \caption{\textbf{spatula} on chair}
        \label{fig:sub5}
    \end{subfigure}
    \hfill
    \begin{subfigure}[b]{0.45\textwidth}
        \centering
        \includegraphics[scale=0.25]{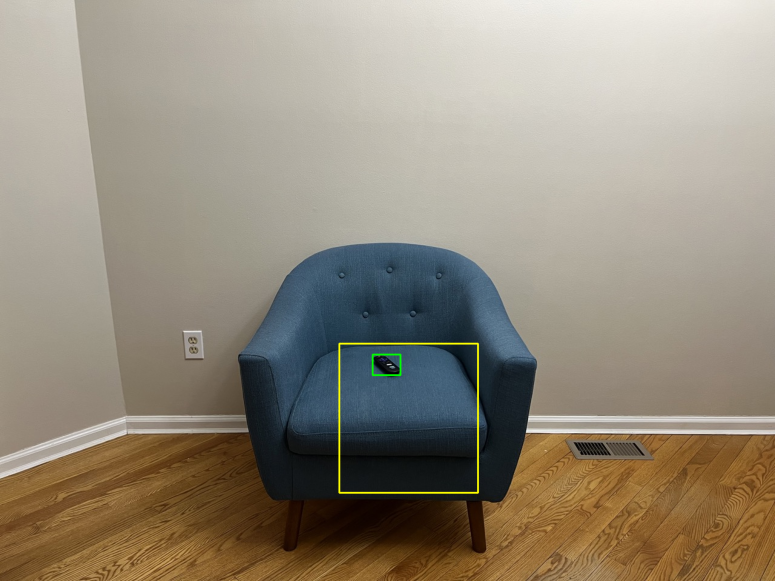}
        \caption{\textbf{remote} on armchair}
        \label{fig:sub6}
    \end{subfigure}

    \begin{subfigure}[b]{0.45\textwidth}
        \centering
        \includegraphics[scale=0.25]{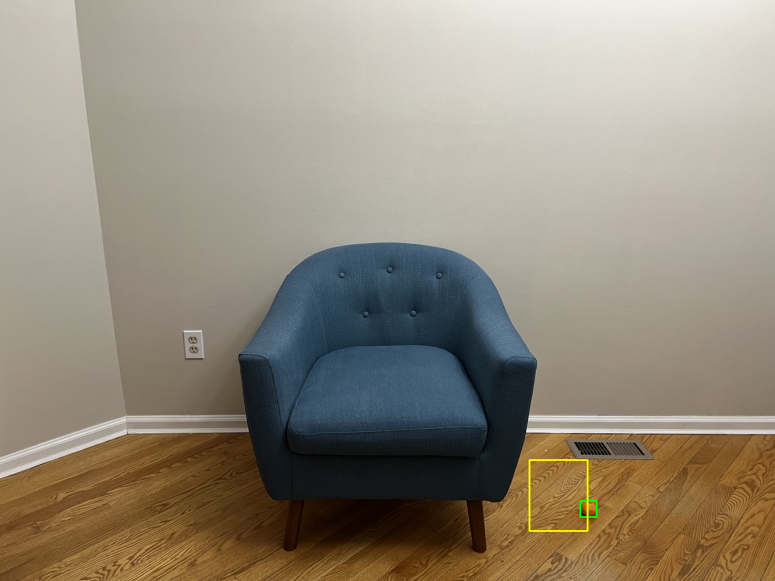}
        \caption{\textbf{orange} right of armchair}
        \label{fig:sub8}
    \end{subfigure}
    \hfill
    \begin{subfigure}[b]{0.45\textwidth}
        \centering
        \includegraphics[scale=0.25]{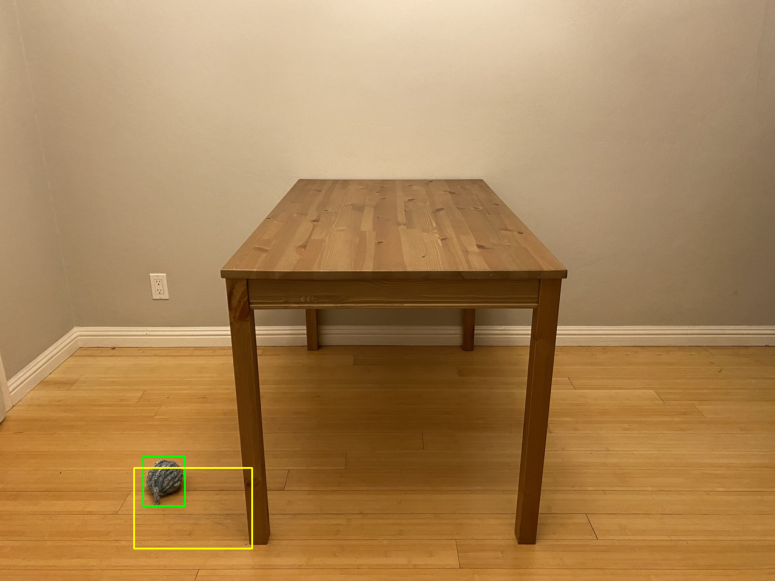}
        \caption{\textbf{ball of yarn} left of table}
        \label{fig:sub7}
    \end{subfigure}

    \begin{subfigure}[b]{0.45\textwidth}
        \centering
        \includegraphics[scale=0.25]{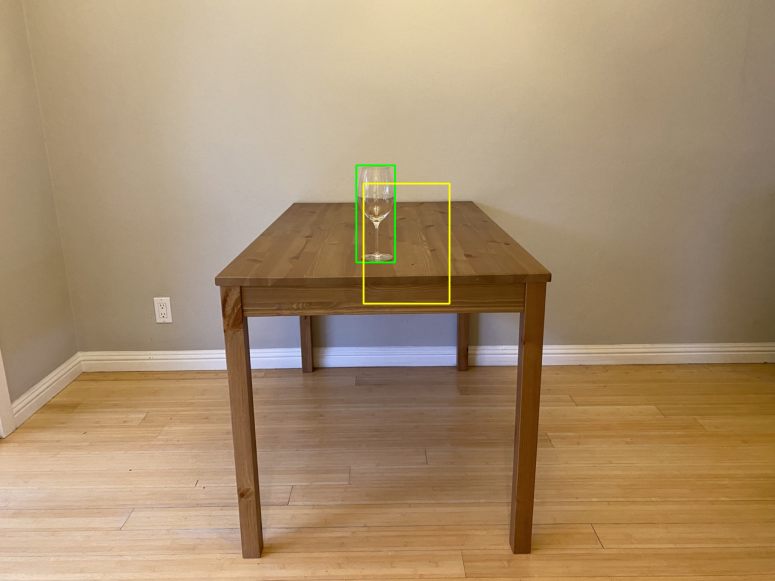}
        \caption{\textbf{wineglass} on table}
        \label{fig:sub9}
    \end{subfigure}
    \hfill
    \begin{subfigure}[b]{0.45\textwidth}
        \centering
        \includegraphics[scale=0.25]{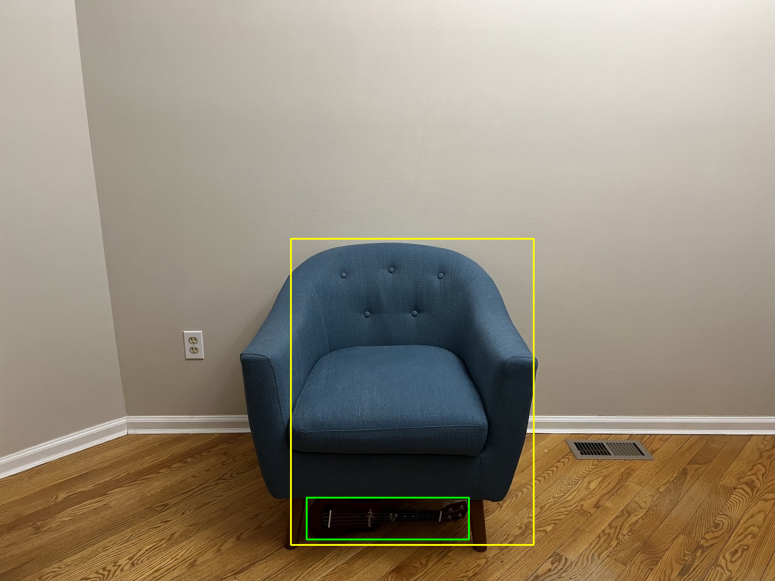}
        \caption{\textbf{banjo} under armchair}
        \label{fig:sub10}
    \end{subfigure}

    \caption{Sample failures in small objects grounding  (i.e., IoU $< 0.5$), which refers to the \textsc{Sub} column results of Subset A in Table \ref{tab:auto_annot_grounding_pipeline}. The pseudo-ground-truth bounding box, which is the GroudningDINO output, is indicated in \colorbox{green}{green}, and the output of \textsc{LLaVA-NeXT-Qwen-1.5-110B}, which is the best-performing MLLM in our grounding/localization experiment, is demonstrated in \colorbox{yellow}{yellow}.}
    \label{fig:groundingerrors}
\end{figure*}

\end{document}